\def\year{2021}\relax
\documentclass[letterpaper]{article}
\usepackage{aaai21} 
\usepackage{times} 
\usepackage{helvet} 
\usepackage{courier} 
\usepackage[hyphens]{url} 
\usepackage{graphicx} 
\usepackage{graphicx}  
\usepackage{natbib}
\usepackage{caption}
\usepackage[utf8]{inputenc}
\usepackage{decision-table}
\usepackage{xspace}
\usepackage{amsthm}
\usepackage{todonotes}
\usepackage{subcaption}
\usepackage{float}

\usepackage{listings}
\theoremstyle{definition}
\newtheorem{example}{Example}[]

\newcommand{\prolog}{Prolog\xspace}
\newcommand{\problog}{ProbLog\xspace}
\newcommand{\pdmn}{pDMN\xspace}

\newcommand{\turnstile}{\hspace{1mm}:\hspace{-1mm}-\hspace{1mm}}
\newcommand{\pnot}{\text{not}\xspace}

\newcommand{\tablesize}{}

\title{A Table-Based Representation for Probabilistic Logic: Preliminary Results}
\author {
    Simon Vandevelde,\textsuperscript{\rm 1,3} 
    Victor Verreet, \textsuperscript{\rm 2,3}
    Luc De Raedt, \textsuperscript{\rm 2,3}
    Joost Vennekens \textsuperscript{\rm 1,3} \\
}
\affiliations {
    \textsuperscript{\rm 1} KU Leuven, De Nayer Campus, Dept. of Computer Science, J.-P.-De Nayerlaan 5, 2860 Sint-Katelijne-Waver, Belgium \\
    \textsuperscript{\rm 2} KU Leuven, Dept. of Computer Science, Celestijnenlaan 200, 3001 Heverlee, Belgium\\
    \textsuperscript{\rm 3} Leuven.AI - KU Leuven Institute for AI, B-3000, Leuven, Belgium \\
    s.vandevelde@kuleuven.be, victor.verreet@kuleuven.be, luc.deraedt@kuleuven.be, joost.vennekens@kuleuven.be
}
\begin{document}



%
%


\maketitle
\begin{abstract}
   We present Probabilistic Decision Model and Notation (\pdmn), a probabilistic extension of Decision Model and Notation (DMN).
   DMN is a modeling notation for deterministic decision logic, which intends to be user-friendly and low in complexity.
   \pdmn extends DMN with probabilistic reasoning, predicates, functions, quantification, and a new hit policy.
   At the same time, it aims to retain DMN's user-friendliness to allow its usage by domain experts without the help of IT staff.
   \pdmn models can be unambigiously translated into \problog programs to answer user queries.
   \problog is a probabilistic extension of \prolog flexible enough to model and reason over any \pdmn model.
\end{abstract}

\section{Introduction}

\problog\cite{ProbLog} is a powerful modeling tool that combines logical reasoning with probabilities.
It supports many inference tasks, such as marginal and conditional probability calculations, allowing it to be used for problems such as Bayesian reasoning and inference in social networks~\cite{gutmann}.
However, \problog rules are often difficult to interpret for domain experts with no familiarity with Probabilistic Logic Programming (PLP).

The Decision Model and Notation standard (DMN) \cite{DMN} is a user-friendly notation for decision logic, published by the Object Management Group (OMG).
The goal of DMN is to be readable and usable by all parties involved in the decision process (business people, IT experts, $\ldots$), as well as being executable.
Currently, it only allows the modelling of deterministic decision processes.

In this paper, we present a preliminary version of Probabilistic Decision Model and Notation (\pdmn): a DMN-like notation for probabilistic logic that aims to combine DMN's intuitive notation with \problog's powerful probabilistic reasoning capabilities.
Our goal is to close the gap between \problog experts and domain experts, by lowering the threshold to understand and interpret probabilistic models, and to possibly allow domain experts to create the models themselves.
This work is similar in spirit to our earlier work on cDMN~\cite{Aerts2020}, that extended DMN with constraint programming.

The contributions of this paper are as follows:
\begin{enumerate}
    \item the \pdmn notation for probabilistic programming, which aims to be user-friendly;
    \item the translation principles of \pdmn into \problog;
    \item an implementation of a \problog-based solver for \pdmn.
\end{enumerate}

This paper is structured as follows.
First, we go over the DMN and ProbLog concepts necessary for this work.
We then introduce our \pdmn syntax, and elaborate on how it differs from standard DMN.
We also present the translation principles of \pdmn into \problog, and very briefly go over our implementation of a \pdmn solver.
Afterwards, we show a full \pdmn implementation of a well-known example, to show \pdmn in action.
Finally, we conclude our preliminary work, and lay out the future work ahead.



\section{Decision Model and Notation}\label{s:dmn}

In DMN, all logic is represented by decision tables.
An example is shown in Fig.~\ref{fig:example_dmn}.
In such a table, the value of the ``output'' variable(s) (in lighter blue) is defined by the value of the ``input'' variable(s) (in darker green).
All variables in DMN are either booleans ($0$-ary predicates), or constants ($0$-ary functions).
Each row of a table represents a decision rule.
A row is applicable if its input values match the actual values of the input variables.
For example, if $\mathit{BMI} = 22$, the second row of the first table is applicable.

The table's \textit{hit policy}, as denoted in the top-left corner, further defines the behavior of the table.
In U(nique) tables, only one row may be applicable for a set of input values.
In A(ny) tables, multiple rows can be applicable, but they must all agree on the output value(s) they assign.
Lastly, in F(irst) tables, the topmost applicable row is always selected.
Besides these so-called \textit{single hit} policies there are also \textit{multiple hit} policies, which are out of the scope of this paper.


A variable which is an output in one table can be an input in another table.
In this way, decisions can be chained together.
For instance, starting from $\mathit{BMI} = 22$, we can first decide that \textit{BMILevel = Normal}  and then use the second table to decide that \textit{Healthy = Yes}.



\begin{figure}
    \tablesize
    \centering
    \dmntable{BMILevel}{U}{BMI}{BMILevel}
         {$< 18.5$, Underweight,
          $[18.5..25]$, Normal,
          $> 25$, Overweight}
    \hspace{1em}
    \dmntable{Healthy}{U}{BMILevel}{Healthy}
         {Normal, Yes,
          {Overweight, Underweight}, No}
    \caption{Example DMN tables}
    \label{fig:example_dmn}
\end{figure}

Another component in DMN, besides the decision tables, is the Decision Requirements Diagram (DRD).
This is a graph that gives an overview of the structure of a DMN model.
However, as this paper focuses in first instance on the decision tables, we will not discuss the DRD further.

\section{Probabilistic Logic Programming}\label{s:problog}

\problog is a probabilistic extension of \prolog.
A \problog program consists of a set of probabilistic facts and a set of \prolog rules. Probabilistic facts are of the form $P_f::f$, with $P_f \in [0, 1]$ a probability and $f$ an atom. The atom $f$ is true or false with probability $P_f$ and $1-P_f$ respectively. Rules are written as $h \turnstile b_1, b_2, \ldots, b_n$ where the atom $h$ is called the head and $b_i$ are the body atoms. The head of a rule may never occur in a probabilistic fact. Whenever all the atoms in the body of a rule are true, the head atom is true as well. A rule can also be annotated with a probability, but this is syntactic sugar for adding a unique atom to the body which is true with the annotated probability. Symbolically, the rule
\begin{equation}
P_r::h \turnstile b_1, b_2, \ldots, b_n
\end{equation}
is translated into
\begin{equation}
h \turnstile b_1, b_2, \ldots, b_n, f_r \quad \text{ and } \quad P_r::f_r
\end{equation}
with $f_r$ a newly created atom. \problog also allows annotated disjunctions (ADs), written as
\begin{equation}
P_1::f_1; P_2::f_2; \ldots; P_n::f_n
\end{equation}
with $\sum_i P_i \leq 1$. An AD denotes a probabilistic choice where every atom $f_i$ is selected to be true with probability $P_i$, but at most one atom in the AD can be selected. If $\sum_i P_i < 1$ it is possible that none are true.

An interpretation is a truth value assignment to every atom occurring in the program. A model of the program is an interpretation that satisfies every rule and follows the closed world assumption. The closed world assumption states that an atom can only be true in a model whenever it can be derived through at least one rule. The probability of any model $M$ of the program is the product of the probabilities of the facts in the model.
The probability of an atom $q$ is the sum of the probabilities of the models in which that atom is true.
Hence,
\begin{equation}
    P(q) = \sum_{M \models q} \prod_{f \in M} P(f)
\end{equation}
where the sum runs over all models $M$ in which $q$ is true and the product runs over all the probabilistic facts $f$ in the model $M$. The probability $P(f)$ is the user given value $P_f$ if $f$ is true in the model $M$, and $1-P_f$ otherwise. An example of a \problog program is given in Example \ref{problog_ex}.

\begin{example} \label{problog_ex}
Consider the program
\begin{equation}
\begin{aligned}
& 0.8 :: a. && c \turnstile a.\\
& 0.3 :: b(1); 0.5 :: b(2); 0.2 :: b(3). && c \turnstile b(1).
\end{aligned}
\end{equation}
where we are interested in the probability of $c$. This program has 6 models, $\{ a, b(1), c \}$, $\{ a, b(2), c \}$, $\{ a, b(3), c \}$, $\{ \pnot(a), b(1), c \}$, $\{ \pnot(a), b(2), \pnot(c) \}$ and $\{ \pnot(a), b(3), \pnot(c) \}$, where $c$ is only true in the first 4 models. Therefore, the probability of $c$ is
\begin{equation}
    P(c) = 0.8 \cdot 0.3 + 0.8 \cdot 0.5 + 0.8 \cdot 0.2 + 0.2 \cdot 0.3 = 0.86
\end{equation}
\end{example}

\section{\pdmn: Syntax}\label{s:syntax}

We now elaborate on the syntax of \pdmn, our DMN extension for probabilistic logic programming.
In \pdmn, there are three types of tables: glossary tables, decision tables, and the query table.

\subsection{Glossary}

Variables in \pdmn, in contrast to standard DMN, are typed $n$-ary functions and predicates.
In order to correctly identify these variables and their arguments, \pdmn introduces three glossary tables in which these should be declared: the \textit{Type} table, the \textit{Predicate} table and the \textit{Function} table.
These glossary tables contain the required meta-information to correctly interpret the \pdmn model.
This is analogous to the approach used in cDMN~\cite{Aerts2020}.

The \textit{Type} table declares the \textit{types} used in a \pdmn model, together with their \textit{domain of elements}.
For example, the \textit{Type} table in Fig.~\ref{fig:example_glossary} declares a type \textit{Person}, which consists of two elements, \textit{ann} and \textit{bob}, and a type \textit{Vaccine}, which consists of the elements \textit{a}, \textit{b} and \textit{n(one)}.

The \textit{Predicate} table declares $n$-ary predicates.
There is no fixed naming syntax for predicates; the arguments of a predicate are those types that appear in its description, and the remaining string is considered the predicate's name.
For example, in the glossary of Fig.~\ref{fig:example_glossary}, \textit{Person is infected} represents a unary predicate \textit{is\_infected}, which denotes for every \textit{Person} (i.e., \textit{ann} and \textit{bob}) whether they are infected.
Similarly, \textit{Person contacted Person} is a binary predicate \textit{contacted} that denotes contact between people.

The \textit{Function} table declares $n$-ary functions.
Analogously to predicates, the function's name contains its arguments.
In contrast to predicates, however, functions map their arguments to the type listed in the \textit{Type} column of the glossary table, instead of to a boolean.
For example, \textit{vaccine of Person} denotes the \textit{Vaccine} for each \textit{Person}, i.e., it maps every person (\textit{ann} and \textit{bob}) to a vaccine (\textit{a, b, n}).

\begin{figure}
    \tablesize
    \centering
    \glossarytable{Type}{Name, Elements}{Person, {ann, bob}, Vaccine, {a, b, n}}
    \hspace{1em}
    \glossarytable{Function}{Name, Type}{vaccine of Person, Vaccine}
    \hspace{1em}
    \glossarytable{Predicate}{Name}{Person is infected, Person contacted Person}
    \caption{Example of a \pdmn glossary}
    \label{fig:example_glossary}
\end{figure}

\subsection{Decision Tables}

\pdmn extends standard DMN decision tables with three new concepts: probabilities, the new \textit{Ch(oice)} hit policy, and quantification.
We will briefly touch on each concept, and show an example.
Firstly, \pdmn allows probabilities in the cells of an output column.
For example, the \textit{h1} and \textit{h2} tables shown in Fig.~\ref{fig:example_coinflip} respectively define a probability of $0.5$ and $0.6$ to flip a coin on its head.
Note that we use \textit{Yes} and \textit{No} to represent \textit{true} and \textit{false} for predicates.
In a table containing probabilities, the output values (such as \textit{Yes}) are not listed in the rules directly, but rather in a separate row above the rules, which contains only output values.
If the conditions of a rule are met, the probability of the output variable taking on a specific value is equal to the value that is listed below this output value in that particular row.



\begin{figure*}
    \begin{subfigure}[b]{\linewidth}
        \tablesize
        \centering
        \pdmnoutputtable{h1}{U}{heads1}{Yes}{0.5}
        \pdmnoutputtable{h2}{U}{heads2}{Yes}{0.6}
        \dmntable{heads}{U}{heads1, heads2}{twoHeads, someHeads}
                           {Yes, Yes, Yes, Yes,
                            Yes, No, No, Yes, 
                            No, Yes, No, Yes,
                            No, No, No, No}
        \caption{Example \pdmn implementation describing two coinflips.}
        \label{fig:example_coinflip}
        \vspace{1em}
    \end{subfigure}
    \begin{subfigure}[b]{\linewidth}
        \tablesize
        \centering
        \pdmntable{Throwing Dice}{Ch}{biased}{die value}{one, two, three, four, five, six}
                 {No, $1/6$, $1/6$, $1/6$, $1/6$, $1/6$, $1/6$,
                  Yes, $0.1$, $0.1$, $0.1$, $0.1$, $0.1$, $0.5$}
        \caption{Example of a pDMN table with the ``Choice'' hit policy.}
        \label{fig:example_choice}
        \vspace{1em}
    \end{subfigure}
    \begin{subfigure}[b]{\linewidth}
        \tablesize
        \centering
        \pdmnoutputtable{Vaccine}{Ch}{vaccine of X}{a, b, n}
        {0.36, 0.63, 0.01}     
        \pdmntable{Infection}{U}{X contacted Y, Y is infected, vaccine of X}{X is infected}{Yes}
         {Yes, Yes, n, 0.8,
          Yes, Yes, a, 0.1,
          Yes, Yes, b, 0.2}
        \caption{Snippet of a \pdmn model implementing infections with vaccination.}
        \label{fig:infection}
    \end{subfigure}
    \caption{Snippets of various \pdmn examples.}
    \label{fig:multifig}
\end{figure*}

The second new concept is the \textit{Ch(oice)} hit policy, which denotes that the output values for the output variable are mutually exclusive (i.e., only one can be assigned to the variable).
This is demonstrated in the table in Fig.~\ref{fig:example_choice}, which states in its first row that an ordinary die has an equal 1/6 chance for any die value, and in its second row that a biased die has a higher chance of resulting in six.
However, because of the \textit{Choice} hit policy, the die can never e.g. be assigned both ``one'' and ``two'' at the same time.
If, for instance, the table had the \textit{Unique} hit policy, it would be possible to have an outcome in which the die has multiple face values at once.


The third and final addition in \pdmn is quantification.
For example, the \textit{Vaccine} table shown in Fig.~\ref{fig:infection} expresses that ``For every Person $X$, there is a chance of 36\% that they have received vaccine $a$, a 63\% chance on vaccine $b$, and a 1\% chance of being unvaccinated.''
The $X$ here represents a quantification variable of type \textit{Person}.
Similarly, the \textit{Infection} table expresses that every person $X$ who had contact with an infected person $Y$ could now also be infected, depending on their vaccine's performance, or lack thereof.

%

\subsection{Query}
The \textit{Query} table is the third type of table present in a \pdmn model, and is used to denote which symbols' probability should be calculated.
Querying the probability of a predicate is done by adding it to the query table, either with specific elements of a type or with a quantification variable.
To query a function, the table should contain a cell of the form $\mathit{func\_name}(\mathit{arg}) = \mathit{val}$.
Here too, it is allowed to write down a specific element of a type or a quantification variable.
Examples of query tables are shown in Fig.~\ref{fig:query}.
The table in Fig.~\ref{subfig:query_a} verifies the probability of flipping two heads and some heads with coins.
Fig.~\ref{subfig:query_b} demonstrates querying predicates with a specific variable value (\textit{bob}), or a quantification variable (\textit{X}).
In the latter case, the probability of the predicate is calculated for every element of the type \textit{Person}.
Lastly, Fig.~\ref{subfig:query_c}, in which we want to know the probability that a die lands on a six, shows the querying syntax for functions.

\begin{figure}
    \centering
    \begin{subfigure}[b]{0.32\linewidth}
        \centering
        \goaltable{Query}{twoHeads, someHeads}
        \caption{}
        \label{subfig:query_a}
    \end{subfigure}
    \begin{subfigure}[b]{0.32\linewidth}
        \centering
        \goaltable{Query}{vaccine of bob, X is infected}
        \caption{}
        \label{subfig:query_b}
    \end{subfigure}
    \begin{subfigure}[b]{0.32\linewidth}
        \centering
        \goaltable{Query}{die value = six}
        \caption{}
        \label{subfig:query_c}
    \end{subfigure}
    \caption{Example \textit{Query} tables.}
    \label{fig:query}
\end{figure}
\section{Translating \pdmn to \problog}\label{s:translation}

To practically use \pdmn models, we translate them into \problog.
We will now go over the general translation principles.
Intuitively, every row of a U-table represents a rule in \problog, with the input variables forming the body, and the output variable forming the head.
If there are multiple output variables present, a rule is created for each of them.
For example, the \textit{heads} table in Fig.~\ref{fig:example_coinflip} translates to the \problog rules shown in (\ref{eq:coins}).
Note that the rows in which the output was \textit{No} are not translated, as these do not need to be explicitly formulated in \problog due to the closed world assumption.

\begin{equation}
\begin{aligned}
    & \mathit{twoHeads} \turnstile \mathit{heads1}, \mathit{heads2}.\\
    & \mathit{someHeads} \turnstile \mathit{heads1}, \mathit{heads2}.\\
    & \mathit{someHeads} \turnstile \mathit{heads1}, \mathit{\pnot(heads2)}.\\
    & \mathit{someHeads} \turnstile \mathit{\pnot(heads1)}, \mathit{heads2}.
\end{aligned}
\label{eq:coins}
\end{equation}

If the output rows of a table contain probabilities, these are added to their respective \problog rules or facts.
E.g., the \textit{h1} table in Fig.~\ref{fig:example_coinflip} translates to the fact $0.5::\mathit{heads1}$.

As explained before, DMN also provides the F(irst) hit policy.
Consider again the \textit{heads} table in Fig.~\ref{fig:example_coinflip}, except we now consider it as an F-table. To translate the first hit behaviour to \problog, for any row in the table we need to add the negation of all the previous rows to the body of the translation. To do this, dummy variables are introduced, representing whether a row has fired or not. The resulting \problog translation for this example is shown in (\ref{eq:first_coins}), where $\mathit{r1}$, $\mathit{r2}$ and $\mathit{r3}$ represent the dummy variables.

\begin{equation}
\begin{aligned}
    & \mathit{r1} \turnstile \mathit{heads1}, \mathit{heads2}.\\
    & \mathit{r2} \turnstile \mathit{heads1}, \mathit{\pnot(heads2)}.\\
    & \mathit{r3} \turnstile \mathit{\pnot(heads1)}, \mathit{heads2}. \\\\
    & \mathit{twoHeads} \turnstile \mathit{r1}.\\
    & \mathit{someHeads} \turnstile \mathit{r1}.\\
    & \mathit{someHeads} \turnstile \mathit{r2}, \mathit{\pnot(r1)}.\\
    & \mathit{someHeads} \turnstile \mathit{r3}, \mathit{\pnot(r1)}, \mathit{\pnot(r2)}.
\end{aligned}
\label{eq:first_coins}
\end{equation}

Tables with the newly introduced \textit{Ch(oice)} hit policy are translated into \problog's annotated disjunctions.
For example, the table shown in Fig.~\ref{fig:example_choice} assigns a value to the 0-ary \textit{die value} function. 
In \problog, $n$-ary functions are represented by an $(n+1)$-ary predicate, resulting in the unary \textit{die\_value} predicate:
\begin{equation}
\begin{aligned}
    & 1/6 :: & \mathit{die\_value}(\mathit{one}); \ldots; 1/6 :: \mathit{die\_value}(\mathit{six}) \\
    & & \turnstile \pnot(\mathit{biased}). \\
    & 1/5 :: & \mathit{die\_value}(\mathit{one}); \ldots; 1/2 :: \mathit{die\_value}(\mathit{six}) \\
    & & \turnstile \mathit{biased}. \\
\end{aligned}
\end{equation}


%

Types declared in the \textit{Type} table in \pdmn are represented by unary predicates in \problog, as the latter is not a typed language.
Additionally, the contents of the \textit{Elements} column are translated into facts.
E.g., \textit{Person}, as shown in the \textit{Type} table in Fig.~\ref{fig:example_glossary}, translates to the facts $\mathit{person}(\mathit{ann})$ and $\mathit{person}(\mathit{bob})$.
When translating a decision table containing quantification, the type of the quantification variable(s) is derived from the glossary, and an atom is added to the \problog rule for each variable to denote its type.
For example, the \pdmn model in Fig.~\ref{fig:infection} translates to the \problog code shown in (\ref{eq:infection}).
Consider e.g. the \textit{Infection} table: it contains two quantification variables, $X$ and $Y$, both of type \textit{Person}.
As such, this is denoted in the \problog rules by adding two atoms to their bodies, $\mathit{person}(X)$ and $\mathit{person}(Y)$, to represent the types of the variables.

\begin{equation}
\begin{aligned}
    &\mathit{person}(\mathit{ann}). \hspace{1em} \mathit{person}(\mathit{bob}). \\
    &\mathit{vaccine}(a). \hspace{1em} \mathit{vaccine}(b). \hspace{1em} \mathit{vaccine}(\mathit{n})\\ 
    & 0.36::\mathit{vaccine}(X, a); 0.63::\mathit{vaccine}(X, b); \\
    & \hspace{3em}0.01::\mathit{vaccine}(X, \mathit{n}) \turnstile person(X).\\
    & 0.8::\mathit{infected}(X) \turnstile \mathit{vaccine}(X, \mathit{n}), \mathit{infected}(Y), \\
    & \hspace{3em} \mathit{contacted}(X, Y), \mathit{person}(X), \mathit{person}(Y).\\
    & 0.1::\mathit{infected}(X) \turnstile \mathit{vaccine}(X, a), \mathit{infected}(Y), \\
    & \hspace{3em} \mathit{contacted}(X, Y), \mathit{person}(X), \mathit{person}(Y).\\
    & 0.2::\mathit{infected}(X) \turnstile \mathit{vaccine}(X, b), \mathit{infected}(Y), \\
    & \hspace{3em} \mathit{contacted}(X, Y), \mathit{person}(X), \mathit{person}(Y).
\end{aligned}
\label{eq:infection}
\end{equation}

The \textit{Query} table is represented in \problog by \textit{query} statements.
For every cell of the table, a new \textit{query} statement is added.
For example, the three query tables shown in Fig.~\ref{fig:query} translate to the following \problog statements:

\begin{equation}
    \begin{aligned}
        & query(twoHeads). \\
        & query(someHeads). \\
       \\ 
        & query(vaccine\_of\_Person(bob)).\\
        & query(person\_is\_infected(X)).\\
       \\ 
        & query(die\_value(six)).
    \end{aligned}
\end{equation}

If no \textit{Query} table is present in a \pdmn model, it is assumed that the probabilities of all symbols of the model should be queried.
In such a case, a \problog \textit{query} rule is generated for every entry in the \textit{Predicate} and \textit{Function} glossary tables.

%
\begin{figure*}[h]
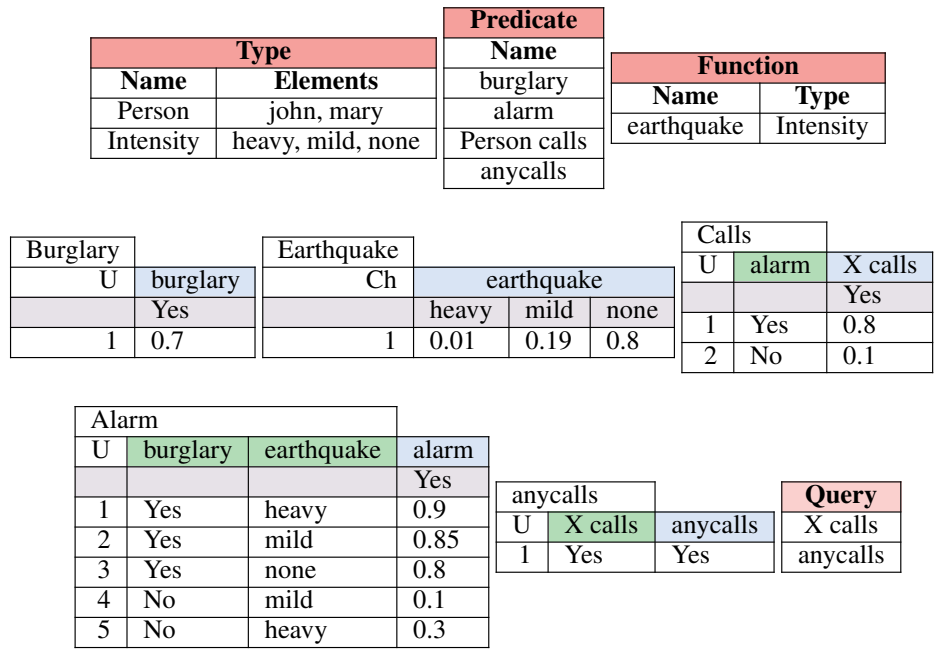

    \centering
    \glossarytable{Type}{Name, Elements}
                  {Person, {john, mary},
                   Intensity, {heavy, mild, none}}
    \glossarytable{Predicate}{Name}
                  {burglary, alarm, Person calls, anycalls}
    \glossarytable{Function}{Name, Type}{earthquake, Intensity}
    
    \pdmnoutputtable{Burglary}{U}{burglary}{Yes}{0.7} 
    \pdmnoutputtable{Earthquake}{Ch}{earthquake}{heavy, mild, none}{0.01, 0.19, 0.8} 
    \pdmntable{Calls}{U}{alarm}{X calls}{Yes}
              {Yes, 0.8,
               No, 0.1}
    \pdmntable{Alarm}{U}{burglary, earthquake}{alarm}{Yes}
              {Yes, heavy, 0.9,
               Yes, mild, 0.85,
               Yes, none, 0.8,
               No, mild, 0.1,
               No, heavy, 0.3}
    \hspace{-1.5em}
    \dmntable{anycalls}{U}{X calls}{anycalls}
             {Yes, Yes}
    \goaltable{Query}{X calls, anycalls}
    \caption{Full \pdmn model for the \textit{Earthquake} example}
    \label{fig:earthquake}
\end{figure*}

\section{Implementation}\label{s:implementation}

To automatically translate \pdmn models to \problog and execute them, the translation principles described earlier have been implemented in a solver\footnote{\url{https://gitlab.com/EAVISE/cdmn/pdmn}}.
This solver is largely based on the solver which we created in earlier work for cDMN~\cite{Aerts2020}, due to the similar nature of the notations.
The input for the solver is a \pdmn model in the form of a \texttt{.xslx} spreadsheet.
Concretely, the solver works in three steps.

First, it interprets all glossary tables in the spreadsheet, beginning with the \textit{Type} table.
For every entry, the solver creates internal \textit{Type} objects, necessary to interpret the arguments used in the \textit{Predicate} and \textit{Function} tables.
The solver then evaluates every decision table one-by-one, using a lex/yacc parser to parse every cell and transform them into a \pdmn expression.
For example, an expression of the form ``\textit{vaccine of bob}'' is translated into ``\textit{vaccine\_of\_person(bob)}''.

Next, all decision tables, are converted into \problog rules in the manner described earlier.
At the same time, the \textit{Query} table is parsed and converted into \problog \textit{query} statements.

Lastly, the generated specification is executed using \problog's Python API, after which the queried probabilities are shown.
In this way, the \pdmn execution process consists of a closed pipeline between \pdmn modelling and \problog execution.

The pDMN solver is available as a Python package, and can be downloaded from its PyPi repository\footnote{\url{https://pypi.org/project/pdmn}}.

\section{Full example}\label{s:example}

In the previous sections, every example only consisted of limited snippets of \pdmn models.
To give a view of what a complete \pdmn model looks like, this section shows a concrete implementation of the well-known \textit{Earthquake} example.
In this example, a house alarm can be triggered by a burglary, by an earthquake of a certain intensity (heavy, mild or none), or by a combination of the two.
Both the burglary and the intensities of the earthquake have a probability associated with them.
If the alarm rings, the neighbours \textit{John} and \textit{Mary} both could either call the home owner, or they could dismiss the alarm as incorrect and ignore it.
We now want to find out the probabilities of either neighbour calling.

The \pdmn model for this example is shown in Fig.~\ref{fig:earthquake}, and consists of the glossary tables, five decision tables and a query table.
In the glossary tables, we first introduce two types, \textit{Person} and \textit{Intensity}, which respectively represent the neighbours and the earthquake intensities.
In the \textit{Predicate} table, we declare four predicates: the 0-ary predicates \textit{burglary}, \textit{alarm} and \textit{anycalls}, and the unary predicate \textit{Person calls}.
To denote the intensity of the earthquake, we make use of the 0-ary function \textit{earthquake}, which will thus either be \textit{heavy}, \textit{mild}, or \textit{none}.

Of the five decision tables, two are straightforwardly used to set the probabilities of a burglary and the earthquake intensities.
As these concepts do not depend on anything, their decision tables contain no input columns.
The \textit{Alarm} table contains a rule for every possible combination of burglary and earthquake to represent the probability of the alarm triggering.
Note that it does not contain a rule in which neither a burglary or an earthquake take place, as the alarm will never trigger in such a situation, thus allowing us to leave out that rule.
The fourth decision table, named \textit{Calls}, expresses that every person $X$ has a certain probability to call the home owner, depending on whether the alarm rings.
Finally, the last decision table defines \textit{anycalls = Yes} whenever any person $X$ calls.

To find the probability of each neighbour calling separately, and the probability of either of them calling, the \textit{Query} table is added to the model in order to finish it.
Translating this model to \problog using the pDMN solver results in the following code:
\allowdisplaybreaks[2]
\begin{align*}
&\%~\mathit{facts}\\
& \mathit{intensity}(\mathit{heavy}). \mathit{intensity}(\mathit{mild}). \mathit{intensity}(\mathit{none}).\\
&\mathit{person}(\mathit{john}). \mathit{person}(\mathit{mary}).\\
&\%~\mathit{Burglary}\\
&0.7::\mathit{burglary}.\\
&\% \mathit{Earthquake}\\
&0.01::\mathit{earthquake}(\mathit{heavy});0.19::\mathit{earthquake}(\mathit{mild});\\
& \hspace{5em}0.8::\mathit{earthquake}(\mathit{none}).\\
&\%~\mathit{Alarm}\\
&0.9::\mathit{alarm} \turnstile \mathit{burglary}, \mathit{earthquake}(\mathit{heavy}).\\
&0.85::\mathit{alarm} \turnstile \mathit{burglary}, \mathit{earthquake}(\mathit{mild}).\\
&0.8::\mathit{alarm} \turnstile \mathit{burglary}, \mathit{earthquake}(\mathit{none}).\\
&0.1::\mathit{alarm} \turnstile \pnot(\mathit{burglary}), \mathit{earthquake}(\mathit{mild}).\\
&0.3::\mathit{alarm} \turnstile \pnot(\mathit{burglary}), \mathit{earthquake}(\mathit{heavy}).\\
&\%~\mathit{Calls}\\
&0.8::\mathit{person\_calls}(\mathit{X}) \turnstile \mathit{alarm}, \mathit{person}(\mathit{X}).\\
&0.1::\mathit{person\_calls}(\mathit{X}) \turnstile \pnot(\mathit{alarm}), \mathit{person}(\mathit{X}).\\
&\%~\mathit{anycalls}\\
&\mathit{anycalls} \turnstile \mathit{person\_calls}(X).\\
&\mathit{query}(\mathit{person\_calls}(\mathit{X})).\\
&\mathit{query}(\mathit{anycalls}).
\end{align*}
              
We can also use the \pdmn solver to execute the example, by running \problog directly.
This results in the following output:
\begin{lstlisting}[frame=single]
>>> pdmn  Examples.xslx  -x  -n  "Earthquake"
{person_calls(mary): 0.501765, person_calls(john): 0.501765,
 anycalls: 0.6319415}
\end{lstlisting}

\section{Conclusion}\label{s:conclusion}

This paper presents a preliminary version of \pdmn, a notation for Probabilistic Logic Programming based on the DMN standard, which aims to combine \problog's expressiveness together with DMN's readability and user-friendliness.
It extends DMN with probabilities, predicates, quantification, and a new hit policy to represent annotated disjunctions.
We lay out the general translation principles of converting \pdmn into \problog code, allowing for the execution of the \pdmn models.
These principles have also been implemented in an automatic conversion tool, which is available for general use.
In future work, we plan on further extending the notation (e.g., with support for more hit policies), formalizing the complete \pdmn semantics, extending the DRD to support probabilities and making a user-friendly interface for the system.


\section{Acknowledgements}
This research received funding from the Flemish Government under the ``Onderzoeksprogramma Artifici\"ele Intelligentie (AI) Vlaanderen'' programme.

\bibliography{biblio}
\end{document}